\documentclass{article}

\usepackage[numbers]{natbib}

 \usepackage[main, final]{neurips_2025}

\usepackage{amsmath}
\usepackage{amssymb}
\usepackage{mathtools}
\usepackage{amsthm}
\usepackage{algorithm}
\usepackage{algorithmic}
\usepackage{multirow}
\usepackage{caption}
\usepackage{subcaption}
\usepackage{blindtext}
\usepackage{placeins}

\theoremstyle{plain}
\newtheorem{theorem}{Theorem}[section]

\newtheorem{lemma}[theorem]{Lemma}

\theoremstyle{definition}

\theoremstyle{remark}

\usepackage[utf8]{inputenc} 
\usepackage[T1]{fontenc}    
\usepackage{hyperref}       
\usepackage{url}            
\usepackage{booktabs}       
\usepackage{amsfonts}       
\usepackage{nicefrac}       
\usepackage{microtype}      
\usepackage{xcolor}         

\DeclareMathOperator{\E}{\mathbb{E}}

\DeclareMathOperator*{\argmin}{arg\ min}
\DeclareMathOperator*{\argmax}{arg\ max}

\newcommand{\bs}[1]{\boldsymbol{#1}}

\title{A Flexible Empirical Bayes Approach to Generalized Linear Models, with Applications to Sparse Logistic Regression}

\author{%
  Dongyue Xie \\
  Department of Statistics\\
  The University of Chicago\\
  Chicago, IL USA \\
  \texttt{dyxie@uchicago.edu} \\
  \And
  Matthew Stephens \\
  Department of Statistics and Department of Human Genetics\\
  The University of Chicago\\
  Chicago, IL USA \\
  \texttt{mstephens@uchicago.edu} \\
}

\begin{document}

\maketitle

\begin{abstract}
We introduce a flexible empirical Bayes approach for fitting Bayesian generalized linear models. Specifically, we adopt a novel mean-field variational inference (VI) method and the prior is estimated within the VI algorithm, making the method tuning-free. Unlike traditional VI methods that optimize the posterior density function, our approach directly optimizes the posterior mean and prior parameters. This formulation reduces the number of parameters to optimize and enables the use of scalable algorithms such as L-BFGS and stochastic gradient descent. Furthermore, our method automatically determines the optimal posterior based on the prior and likelihood, distinguishing it from existing VI methods that often assume a Gaussian variational. Our approach represents a unified framework applicable to a wide range of exponential family distributions, removing the need to develop unique VI methods for each combination of likelihood and prior distributions. We apply the framework to solve sparse logistic regression and demonstrate the superior predictive performance of our method in extensive numerical studies, by comparing it to prevalent sparse logistic regression approaches. 
\end{abstract}

\section{Introduction}
\label{sec:intro}

The generalized linear model (GLM), an extension of the Gaussian linear model to the exponential family, has been widely used in various applications. One of the most notable  examples is the logistic regression model, which is particularly  prevalent in statistical learning for classification problems.
Bayesian methods for GLM have also been extensively studied \cite{maity2019integration,makalic2016high,narisetty2018skinny,yi2019bhglm,zhang2019novel}. These methods typically assume priors on the regression coefficients and conduct posterior inference via sampling. 

In high-dimensional settings, regularization techniques are commonly applied to improve model generalization and to address issues related to the stability of the solution \cite{hastie2009elements}. 
Existing methods for penalized GLM typically apply penalties to the coefficients to encourage sparsity (near or equal to 0), such as lasso ($l_1$, \cite{tibshirani1996regression}), ridge ($l_2$, \cite{hoerl1970ridge}), elastic net \cite{zou2005regularization}, Minimax Concave Penalty (MCP, \cite{zhang2010nearly}), and Smoothly Clipped Absolute Deviation (SCAD, \cite{fan2001variable}). 
 Likewise, many efficient algorithms and software packages have been developed for these methods \cite{bertrand2022beyond,breheny2011coordinate,carmichael2021yaglm,friedman2010regularization,liu2022fast}.  
To induce shrinkage and sparsity on the regression coefficients in  Bayesian methods, sparsity-inducing priors are used such as spike-and-slab type priors \cite{chipman1996bayesian,george1993variable,Ishwaran2005,mitchell1988bayesian,ray2020spike} and continuous shrinkage priors such as horseshoe \citep{carvalho2009handling,carvalho2010horseshoe}, horseshoe $+$ \citep{horseshoeplus}, R2D2 \citep{zhang2022bayesian, aguilar2023intuitive}, which have proven effective in handling high-dimensional data analysis. In this paper, we will use the terms penalized GLM and sparse GLM interchangeably to denote shrinkage estimation of the regression coefficients (towards 0).


There are some limitations to the aforementioned methods, primarily related to selecting the most appropriate regularization and computations. For penalized GLM, existing methods typically pre-choose the penalties before model fitting. However, the pre-chosen penalty might not be optimal for the dataset. Furthermore, the penalty level is controlled by additional parameters that often require tuning via cross-validation (CV). If there are multiple tuning parameters (or multiple penalties to compare), the entire procedure can be computationally demanding. 
On the other hand, 
fast Bayesian GLM are developed using variational inference methods, but they are only suitable for certain priors and likelihoods. For example, varbvs \cite{carbonetto2012scalable} and sparsevb \cite{ray2020spike} use the Jordan-Jaakkola lower bound \cite{jaakkola1997variational,jordan1999introduction} for Bayesian logistic regression with a spike-and-slab prior; a lower bound for Poisson likelihood was developed by \citet{seeger2012fast} based on the softplus link function; \citet{arridge2018variational} developed variational Gaussian approximation for Poisson regression. These methods are specific to a single distribution, and do not apply to others. \citet{wang2013variational} introduced a class of variational inference method for nonconjugate models based on Laplace and delta methods but are limited to Gaussian posterior distribution on the nonconjugate variable. 

In this work, we develop a flexible and unified framework, the empirical Bayes generalized linear model (EBGLM), for Bayesian GLM inference that addresses all the aforementioned concerns.
In particular, we adopt an empirical Bayes (EB) approach combined with variational inference. Our framework is inspired by the penalty formulation of an empirical Bayes multiple linear regression with Gaussian error, introduced by \citet{kim2022flexible}. While the EBGLM also utilizes a penalty formulation, our method is more general than \citet{kim2022flexible} in both problem formulation and applications. We highlight the main advantages and contributions of the proposed method:
\begin{itemize}
    \item[1.] The EBGLM framework is compatible with data distributions within the exponential family and a wide range of prior choices.
    \item[2.] By using the EB approach, the optimal parameters in the prior are estimated through an optimization problem, making the method self-tuning. 
    \item[3.]The optimal variational posterior is automatically determined without pre-specification, due to its strong connection with the Bayesian normal means problem \cite{willwerscheid2021ebnm}.
    \item[4.] The objective function has a closed-form expression and can be evaluated explicitly, unlike many VI methods that require stochastic gradient estimators and Monte Carlo sampling.
    \item[5.] The VI method supports direct use of well-developed optimization routines such as L-BFGS \cite{liu1989limited} and gradient descent \cite{boyd2004convex}, similar in spirit to recent advances in variational inference such as stochastic variational inference \citep{hoffman2013stochastic} and automatic differentiation VI \cite{kucukelbir2017automatic}.
\end{itemize}
 We demonstrate the framework by solving sparse high-dimensional logistic regression and show the outstanding predictive performance of our method through extensive numerical studies, comparing it to prevalent sparse logistic regression approaches.

 \section{Empirical Bayes GLM} 
Given a feature vector $\bs x\in \mathbb{R}^{p}$, a GLM predicts the response $y\in\mathbb{R}$ using a linear predictor $\bs x^T\bs \beta$, where $\bs\beta\in\mathbb{R}^{p}$ is a $p$-dimensional regression coefficient vector.
The two most common examples of GLMs are Gaussian linear model, where $y|\bs \beta\sim N(\bs x^T\bs \beta,\sigma^2)$, and the logistic regression model, where $y|\bs \beta\sim \text{Bernoulli}(\sigma(\bs x^T\bs \beta))$, with $\sigma(\eta) = \exp(\eta)/(1+\exp(\eta))$ being the sigmoid function. In a Bayesian setting, we place a prior on $\bs\beta$ and study the posterior distribution. This paper addresses the estimation of the regression coefficients $\bs{\beta}$ by computing the posterior mean and using it for prediction.

Given data $\{(\bs x_i, y_i)\}_{i=1}^{n}$, we consider a canonical GLM of the form:  
\begin{equation}\label{eqn:glm}
    p(y_i;\bs\beta,\phi) = \exp\left(\frac{y_i\eta_i - b(\eta_i)}{a(\phi)} + c(y_i,\phi)\right),
\end{equation}
where $\eta_i = \bs x_i^T\bs\beta$ is the linear predictor and is known as the natural or canonical parameter in the exponential family. The dispersion parameter $\phi$ is assumed to be known in this paper, so we write $p(y_i;\bs\beta,\phi) = p(y_i\mid\bs\beta)$ for simplicity. The functions $a(\cdot)$, $b(\cdot)$, and $c(\cdot)$ vary across different models. For example, for logistic regression, $a(\phi)=1$, $b(\eta) = \log(1+\exp(\eta))$, $c(y_i,\phi) = 0$. Therefore $p(y_i\mid\bs\beta) = \exp\left(y_i\eta_i - \log(1+\exp(\eta_i))\right)$. Refer to Table 2.1 in \citet{mccullagh2019generalized} for other distributions.

We consider a prior on $\beta_j$ for $j=1,2,...,p$: 
\begin{equation*}
    \beta_j\overset{\text{i.i.d}}{\sim} g(\cdot), g\in\mathcal{G},
\end{equation*}
where $\mathcal{G}$ can be any arbitrary family of prior distributions. Note that if $x_{i1} = 1, \forall i = 1, \dots, n$, and $\beta_1$ represents the intercept,  it is common practice to use a flat (non-informative) prior, or no prior, on the intercept. In this paper, we adopt the latter approach.
The standard empirical Bayes procedure first estimates the prior $g$ by maximizing the log marginal likelihood 
$l(g) = \log p(\bs y;g)=\log\int p(\bs y|\bs\beta)g(\bs\beta)d\bs\beta,$
where $\bs y = (y_1, \dots, y_{n})$,
then calculates the posterior distribution 
$p(\bs\beta|\bs y,\hat g)$
conditional on the estimated prior $\hat g$. With the posterior distribution $p(\bs\beta|\bs y,\hat g)$, we can obtain point estimators for $\bs\beta$, such as posterior mean or median,  as well as conduct uncertainty quantification. However, in the settings here, both steps are non-trivial, especially since we aim to develop a general framework that accommodates a wide range of prior distributions and likelihoods. Therefore, for prior estimation and posterior inference, we opt for the variational inference approach. Before presenting our novel framework, we first provide a brief review of variational inference.

\subsection{Review of variational inference}

Variational Inference (VI) turns the Bayesian posterior inference problem into an optimization problem by approximating the true posterior $p(\bs\beta|\bs y;g)$ with a simpler, more tractable distribution $q(\bs\beta)$. A canonical review of VI is given by \citet{blei2017variational}. Specifically, VI finds 
\begin{equation}\label{eqn:VI_KL}
    q^*(\bs\beta) = \argmin_{q\in\mathcal{Q}} D_{KL}(q(\bs\beta)\Vert p(\bs\beta|\bs y;g)),
\end{equation}
where prior $g$ is given, $\mathcal{Q}$ is a family of approximate densities and $D_{KL}$ is the Kullback-Leibler (KL) divergence. 

Directly minimizing the KL divergence \eqref{eqn:VI_KL} is intractable because the true posterior $p(\bs{\beta}|\bs{y};g)$ is unknown, and so instead the problem is tackled by minimizing the following Evidence Lower Bound (ELBO), which is a lower bound on the log marginal likelihood $\log p(\bs{y};g)$:
\begin{equation*}
\begin{split}
        F(q, g;\bs y) &=\log p(\bs y;g) - D_{KL}(q(\bs\beta)\Vert p(\bs\beta|\bs y;g)),
        \\
        &=  \mathbb{E}_{q(\bs{\beta})}(\log p(\bs{y}, \bs{\beta};g) - \log q(\bs{\beta})) .
\end{split}
\end{equation*}

Variational Empirical Bayes (VEB) integrates variational inference and empirical Bayes into a single optimization problem:
\begin{equation}
\label{eqn:veb_obj}
    q^*(\bs\beta),\hat g = \argmax_{q\in\mathcal{Q}, g\in\mathcal{G}} F(q,g;\bs y).
\end{equation}

\subsection{A novel MFVI formulation}

For the EBGLM model, we take the MFVI approach and assume the approximate posterior factorizes as
\begin{equation*}
    q(\bs\beta) = \prod_j q_{j}(\beta_j),
\end{equation*}
where $q_{j}(\cdot)$ is the density for a univariate distribution. Under the exponential family distribution \eqref{eqn:glm}, we define
$l(\eta_i) = (y_i\eta_i-b(\eta_i))/a(\phi) = \log p(y_i|\bs\beta) + \textnormal{const},$
where $\eta_i = \bs x_i^T\bs\beta$. The corresponding ELBO for EBGLM is given by:
\begin{equation}
\label{eqn:elbo_glm}
    \begin{split}
        F(q,g) 
        &= \E_{q}\sum_{i=1}^n l(\eta_i) - \sum_{j=1}^p D_{KL}(q_{j}|| g).
    \end{split}
\end{equation}

  Maximizing $F(q,g)$ is still challenging since the optimization variables are distributions, and the expectations in the ELBO often lack an explicit form.   In the rest of this section, we introduce a novel approach that defines an approximate ELBO, and transforms it into a function of real-valued parameters -- specifically, the posterior means and prior parameters. The new objective function has an analytic form, and so can be optimized using any existing optimization methods such as gradient-descent.  Moreover, the optimal variational posterior is determined automatically through a Bayesian normal mean problem. 
  
  These contrast with a large body of work on black-box type of variational inference methods, which typically assume either specific parametric forms for the variational posterior or employ more flexible, general approximations, and address intractable expectations by employing stochastic gradient estimators and Monte Carlo sampling. For instance,  black box variational inference (BBVI, \citet{ranganath2014black}) estimates gradients using the score function estimator but it requires variance reduction techniques such as control variates. Automatic differentiation variational inference (ADVI, \citet{kucukelbir2017automatic}) restricts the variational family to mean-field or full-rank Gaussians and uses the reparameterization trick to obtain low-variance gradients.  Other notable advancements include Pathfinder \citep{zhang2022pathfinder} and stochastic natural gradient variational inference (NGVI, \citet{wu2024understanding}). 



\subsubsection{Approximate ELBO}
Let $\bar{\bs\beta}_q = \E_{q}(\bs\beta)$ denote the mean of $\bs\beta$ under the variational posterior distribution $q(\cdot)$. We consider the Taylor expansion of $l(\eta_i)$ around the approximate posterior mean linear predictor $\bar\eta_i= \bs x_i^T\bar{\bs\beta}_q$:  
\begin{equation}
\begin{split}
        \tilde l(\eta_i) &= l(\bar\eta_i) + l'(\bar\eta_i)(\eta_i-\bar\eta_i) + \frac{1}{2}l''(\bar\eta_i)(\eta_i-\bar\eta_i)^2.
\end{split}
\end{equation} 
By replacing $l(\eta_i)$ with $\tilde l(\eta_i)$ in \eqref{eqn:elbo_glm}, the approximate ELBO is then
\begin{equation}\label{eqn:approx_elbo}
    \begin{split}
        \tilde F(q,g) &= \E_{q}\sum_i \tilde l(\eta_i) - \sum_j D_{KL}(q_{j}|| g).
    \end{split}
\end{equation} 

We note that the delta method for VI in non-conjugate models, proposed by \citet{wang2013variational}, also employs a second-order Taylor series expansion around the mean of the variational posterior distribution. However, the delta method assumes a multivariate Gaussian variational posterior and applies the Taylor expansion to the log joint density (i.e., the sum of the log-likelihood and the log prior).  In contrast, our method applies the expansion to the log-likelihood, and as we will demonstrate, the optimal variational posterior is determined automatically.

The approximate ELBO function $\tilde F(q,g)$ remains a function of the distributions $q(\cdot)$. The following Lemma connects the VEB approach to a penalized regression framework by reformulating the optimization problem in terms of real-valued parameters instead of the posterior distributions.
\begin{lemma}\label{lemma:h}
   Define 
   \[r_j( \bs\theta,g)=\min_{q_{j}:\E_{q_{j}}(\beta_j)=\theta_j}\left(\frac{1}{2s^2_j(\bs\theta)} V_{q_{j}} + D_{KL}(q_{j}||g)\right),\] 
  where  $V_{q_{j}}$ is the posterior variance of $\beta_j$ under distribution $q_j$ and 
$s^2_j(\bs\theta) = a(\phi)/(\sum_i b''(\bs x_i^T\bs\theta)x_{ij}^2)$. 
  Then we have
  \[\max_{q,g} \tilde F(q,g) = \min_{\bs\theta,g}h(\bs\theta,g),\]
  where \begin{equation*}
    h(\bs\theta,g)=-\sum_i l(\bs x_i^T\bs\theta) + \sum_j r_j(\bs\theta,g).
\end{equation*}
\end{lemma}



Lemma \ref{lemma:h} shows that maximizing the approximate ELBO $\tilde{F}(q, g)$ can be achieved by minimizing the real-valued objective function $h(\bs\theta,g)$. According to the definition of $r_{j}$, the objective function $h(\bs\theta,g)$ depends on the posterior mean $\bs\theta$ and prior (parameters) $g$. Minimizing $h(\bs\theta,g)$ directly yields the posterior mean and the estimated prior. In particular, $h(\bs\theta,g)$ resembles commonly used forms in statistical problems, such as penalized regression, comprising a summation of the log-likelihood $l(\cdot)$ and a penalty term $r(\cdot)$. However, we emphasize that this penalty applies to the posterior mean, which is the Bayes risk optimal estimator under squared loss. This distinguishes our method from the well-known maximum a posteriori (MAP) estimator in Bayesian statistics, which also incorporates a penalization formulation but targets the posterior mode, with a much simpler penalty. Specifically, the posterior mode might not be an ideal estimator in sparse regression settings; for example, if the prior includes a point mass at 0, as seen in many spike-and-slab priors, the posterior mode will trivially be at 0. 

\subsubsection{Evaluating approximate ELBO}
Evaluation of $h(\bs\theta,g)$ involves solving $r_{j}$ for each $j = 1, \dots p$.
We will show that the optimal $q_j$ in $r_j(\bs\theta, g)$ 
is a convolution of a Gaussian density and prior $g$. 

We first introduce the univariate Bayesian normal mean (BNM) inference model as follows: 
\begin{equation}
    \begin{split}
        z|b&\sim N(b,s^2),
        \\
        b&\sim g(\cdot),
    \end{split}
\end{equation}
where $g$ is the prior and $s^2$ is the variance. The log marginal likelihood in BNM model is
\begin{equation}\label{eq:logmargin}
    l_{\text{NM}}(z;g, s^2) = \log\int N(z;b,s^2)g(b)db.
\end{equation}

\begin{theorem}\label{thm:penalty}
With the same definitions as in Lemma \ref{lemma:h}
\begin{equation*} 
\begin{split}
        r_j(\bs\theta,g) 
    =& -l_{\text{NM}}(z_g(\theta_j);g,s^2_j(\bs\theta)) \\& \quad+ \log N(z_g(\theta_j);\theta_j,s_j^2(\bs\theta)),
\end{split}
\end{equation*}
 where $l_{\text{NM}}(\cdot)$ is defined in  ~\eqref{eq:logmargin}  and $z_g(\theta_j)$ is found by solving for $z$ in the following root finding problem for each $j$:
\begin{equation}\label{eqn:root_finding}
    \theta_j = z + s_j^2(\bs\theta)l'_{\text{NM}}(z;g,s_j^2(\bs\theta)),
\end{equation}
where $l'_{\text{NM}}(\cdot)$ is the derivative of $l_{\text{NM}}(\cdot)$ with respect to $z$. Moreover, the optimal $q_j$ in $r_j(\bs\theta,g)$ is  
\begin{equation*}
        q_j(\cdot) = \frac{g(\cdot)N(\cdot; z_g(\theta_j),s_j^2(\bs\theta))}{c(z_g(\theta_j))},
    \end{equation*}
    where $c(z_g(\theta_j)) = \int g(b)N(z_g(\theta_j);b,s_j^2(\bs\theta)) db$, the log marginal likelihood in a BNM model. 
    
\end{theorem}

 In Theorem \ref{thm:penalty}, $z_g(\theta_j)$ is obtained by solving an ``inverse'' BNM problem using Tweedie's formula \cite{efron2011tweedie}, which involves finding the observations that result in a given posterior mean and prior.  We demonstrate that the optimal variational posterior approximation, under the approximate ELBO, is the posterior distribution of the BNM problem. This distinguishes our method from existing ones that rely on Gaussian variational posterior distributions. To evaluate $h(\bs\theta,g)$, we only require the log marginal likelihood and its derivative in a BNM problem (to solve for $z_g(\theta_j)$). This approach enables us to  work directly with the marginal likelihood, even without a fully-specified prior distribution. For further discussion on empirical Bayes $g$-modelling and $f$-modelling, see \citet{efron2014}.

Finally, we solve the following optimization to obtain the posterior mean $\bs\theta$, and estimated prior $\hat g$ for EBGLM:  
\begin{equation}\label{eqn:pen_obj1}
    \begin{split}
        \min_{\bs\theta,g} h(\bs\theta,g) = - \sum_i l(\bs x_i^T\bs\theta) + \sum_j r_j(\bs\theta,g).
    \end{split}
\end{equation} 
To obtain the full variational posterior distribution $q_j(\cdot)$, one can plug the ``data'' $z_{\hat g}(\theta_j)$ for $j=1,2,...,p$ and $\hat g$ into the BNM \citep{willwerscheid2021ebnm} problem. We demonstrate the effectiveness of the proposed VI method on examples shown in Appendix~\ref{sec:illustration_vi}. The method accurately recovers the posterior mean across all settings, and provides a good approximation to the posterior distribution when the correlations among features are weak.



 \section{Choice of priors for sparse GLM}\label{sec:priors}

In this section, we consider the sparse GLM problem using our novel framework. For the shrinkage estimation of regression coefficients, we employ the spike-and-slab type priors. These priors are widely applied in various statistical applications, including variable selection and false discovery rate control \cite{mitchell1988bayesian,biswas2022scalable,hernandez2013generalized,Ishwaran_Rao_2005,stephens2017false}, as well as shrinkage estimation in linear regression and non-parametric regression \cite{bai2022spike,piironen2017sparsity,ray2020spike,xing2021flexible}. 
In this paper we consider three priors known for their widespread use in statistical modeling:
\begin{itemize}
    \item Point Normal prior: $\beta_j\sim \pi_0 \delta_0 + (1-\pi_0) N(0,\sigma^2),$ where $\delta_0$ denotes a Dirac delta distribution (point-mass) at 0, and $\{\pi_0,\sigma^2\}$ are parameters to be estimated from data. 
    \item Point Laplace prior: $\beta_j\sim \pi_0 \delta_0 + (1-\pi_0) \text{Laplace}(0,\sigma^2),$ using the same notations as in the Point Normal prior example. The corresponding marginal distribution in the BNM problem is detailed in Appendix \ref{appendix:point_laplace}.
    \item Adaptive shrinkage (ash) prior \cite{stephens2017false}: $\beta_j\sim \sum_{k=1}^K\pi_k N(0,\sigma^2_k),$ where ${\pi_k}$'s are parameters to be estimated from data. The number of clusters $K$, and variances $\sigma^2_k$ (a dense grid of variances spanning from very small to very large values) are known. 
\end{itemize}

In our empirical Bayes framework, the optimal parameters for each prior class are estimated from the data. As stated in Theorem \ref{thm:penalty}, evaluating the ELBO requires only the log marginal likelihood and its derivative from the corresponding BNM problem. The optimal posterior for regression coefficients also corresponds to the posterior in the BNM problem. For point normal and ash priors, both the marginal and posterior distributions are mixtures of normals. Details for the point Laplace distribution are provided in Appendix \ref{appendix:point_laplace}. We will use all three of these priors in the simulations in Section \ref{section:simu}. For the ash prior, we set $K=20$, $\sigma^2_0 = 0, \sigma^2_1 = 0.01$, and the maximum $\sigma^2_K=n$. The remaining variances are $\sigma^2_k = \sigma^2_1 \times a^{k-1}$, where $a = (\sigma^2_K/\sigma^2_1)^{1/(K-1)}$. 

When studying penalty functions, we can visualize the penalties (see in Figure \ref{fig:other_penalty}) to understand the sparsity property of the estimate. Similarly, we can visualize the shrinkage effect of the priors through our penalization formulation. 
In Figure \ref{fig:pn_penalty}, we visualize  the varying shape of the penalty function $r(\cdot)$ by considering different combination of $\pi_0$ and $\sigma^2$ in the point normal prior when $X=I_{n\times n}$ in logistic regression. 
The flexibility of the point normal prior is obvious that different combinations of $\{\pi_0,\sigma^2\}$ can lead to varied penalties. More importantly, the optimal penalty shape is data-driven due to the empirical Bayes approach. 

\section{Simulations}
\label{section:simu}
We apply the EBGLM framework to sparse logistic regression, which is a widely used GLM in applications such as fraud detection and click-through rate prediction.  The density function is
$p(y_i;\bs\beta) = \exp\left(y_i(\bs x_i^{T}\bs\beta) - \log(1+\exp(\bs x_i^{T}\bs\beta))\right).$ 

We compare 
our method against other well-established sparse logistic regression methods, including 
\begin{itemize}
    \item Lasso (\texttt{glmnet, \citet{tibshirani1996regression}} package in \texttt{R}): Applies an $l_1$ penalty to encourage sparsity by shrinking some coefficients to zero.
    \item Ridge (\texttt{glmnet}, \citet{hoerl1970ridge}): Uses an $l_2$ penalty to shrink coefficients.
\item Elastic net (\texttt{glmnet}, \citet{zou2005regularization}): Combines $l_1$ and $l_2$ penalties.
\item SCAD (\texttt{ncvreg} package in \texttt{R}, \citet{fan2001variable}): Employs a nonconvex penalty that aims to reduce bias in large coefficients while maintaining sparsity.
\item MAD (\texttt{ncvreg}, \citet{zhang2010nearly}): Uses a nonconvex penalty with  smoothing properties.
\item L0Learn (\texttt{Python}, \citet{hazimeh2023l0learn}): Directly approximates an $l_0$ penalty, allowing explicit control of the number of nonzero coefficients.
\item varbvs (\texttt{R}, \citet{carbonetto2012scalable}): Uses a spike-and-slab (point normal) prior with variational inference for varaible selection in logistic regression.
\item sparsevb (\texttt{R}, \citet{ray2020spike}): Uses variational Bayes with Laplace slabs in  logistic regression.
\end{itemize}
For methods requiring parameter tuning, we employed 10-fold CV by default. For others, we adhered to the default settings provided by the software packages. 
We chose the optimal model for lasso and ridge based on the 1-standard error rule as recommended by \citet{breiman2017classification,friedman2010regularization}. Although the elastic net does not have built-in CV support, we developed a custom CV procedure with $\alpha$ ranging from 0 to 1 with step being 0.1. 
For the EBGLM method, we consider three different priors introduced in Section \ref{sec:priors}: EBGLM with a point normal prior (EBGLM-pn), a point Laplace prior (EBGLM-pl), and an ash prior (EBGLM-ash). 
We optimized the EBGLM objective function using the L-BFGS-B optimizer \cite{byrd1995limited}. The step size is determined by a backtracking line search, and the memory size of L-BFGS is $\mathcal{O}(mp)$ where it stores the $m$ most recent updates and we used $m=10$ as it's a common default choice in practice. We used the classical bisection method to solve the root-finding problem in \eqref{eqn:root_finding}. The prior parameters and posterior mean are initialized at the cross-validated lasso results using the ``1se'' rule. The EBGLM \texttt{Python} package and code for reproducing the results can be found at \url{https://github.com/DongyueXie/vebglm-paper}. All experiments were conducted on a 
Linux system with an i9-10900 processor and 16GB memory.

In the simulation, we consider different settings with varying sample sizes $n$, dimensionalities $p$, sparsity levels $s$, correlations $\rho$ among the columns of the design matrix, and distributions for generating regression coefficients. Specifically, we try
\begin{enumerate}
    \item Sample size: $n\in\{200,300,500,750,1000,1500,2000,3000\}$.
    \item Correlation among features: $\rho\in\{0,0.1,0.2,...,0.9,0.95,0.99\}$.
    \item Number of non-zero coefficients (sparsity): $s\in\{1,5,10,30,50,100,200,300\}$.
    \item Dimensions: $p\in\{20,50,100,200,500,1000,2000,3000\}$.
    \item Distributions of coefficients: standard normal, standard Laplace, 
    constant (value equals $1$), Uniform$[-1,1]$.
\end{enumerate}
The default simulation configuration is $n=500, p=1000, s=20, \rho=0$, and non-zero coefficients are drawn from a standard normal distribution. For each setting, we 
train models and test them on $5000$ testing samples. We evaluated performance using the mean area under the curve (AUC) of the receiver operating characteristic (ROC), averaged over $100$ replications.

\begin{figure*}[htbp]
    \centering
    \includegraphics[width=0.49\textwidth]{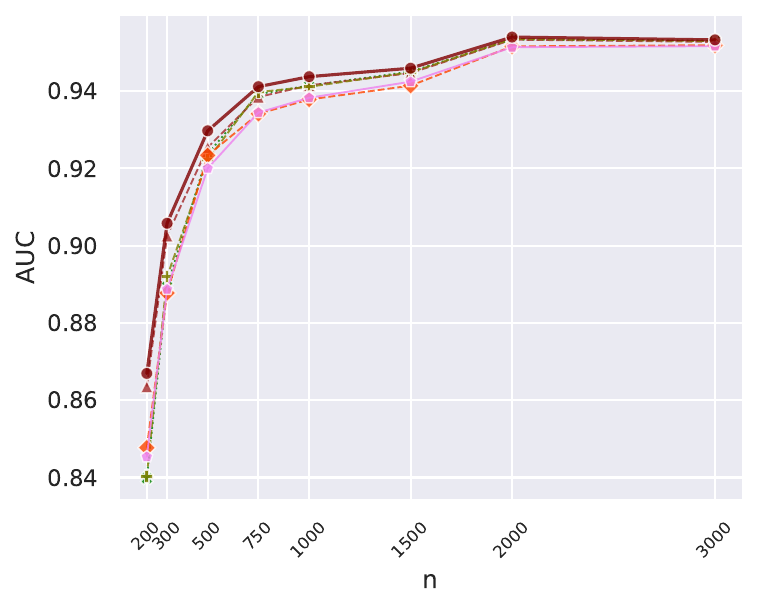}
    \includegraphics[width=0.49\textwidth]{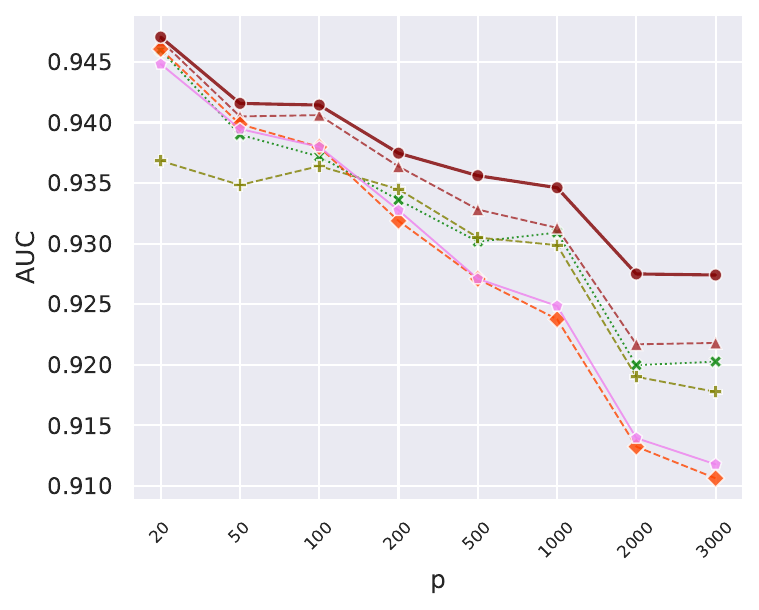}
    \includegraphics[width=0.49\textwidth]{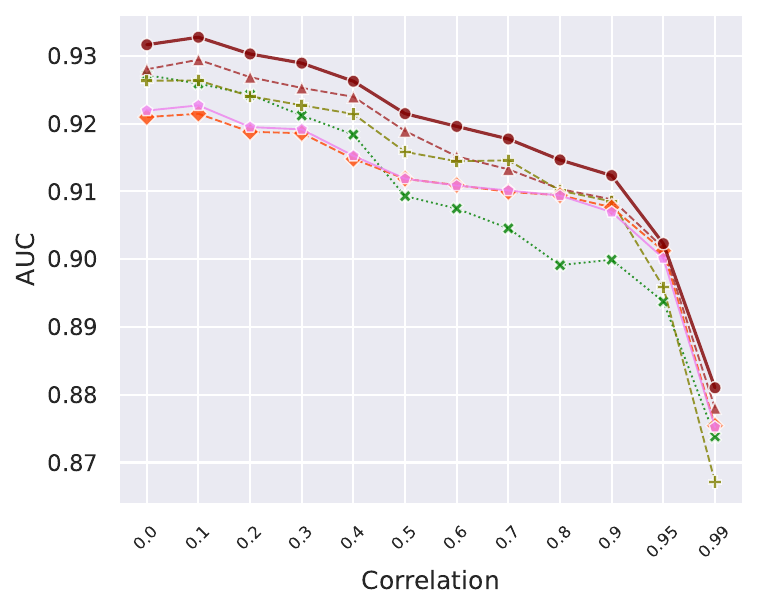}
    \includegraphics[width=0.49\textwidth]{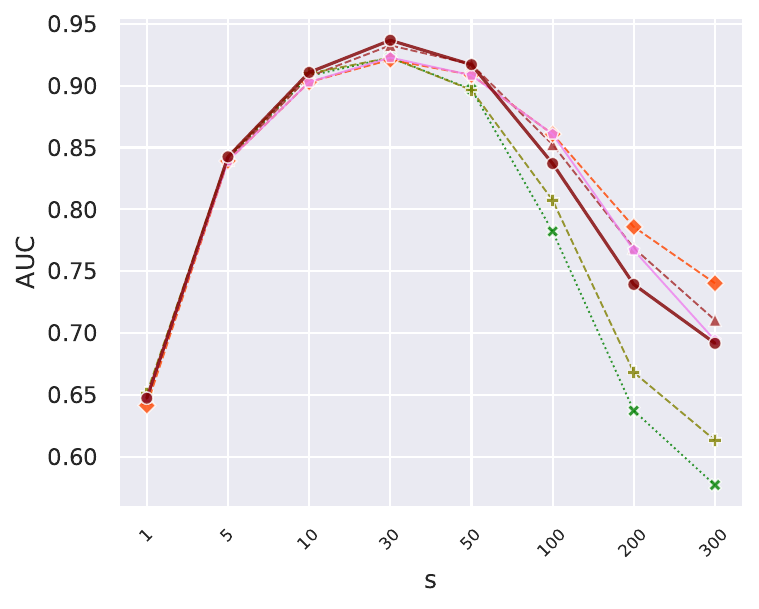}
    \includegraphics[width=0.51\textwidth]{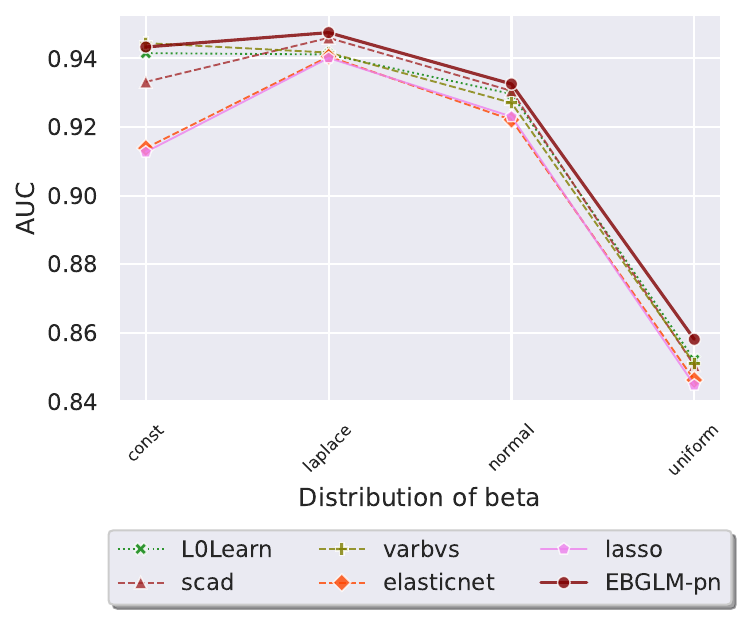}
    \caption{\textbf{Average AUC across five simulation settings}. 
        We vary \textit{(top-left)} sample size \(n\), 
        \textit{(top-right)} dimensionality \(p\), 
        \textit{(middle-left)} correlation \(\rho\) among features, 
        \textit{(middle-right)} sparsity level \(s\) (number of nonzero coefficients), 
        and \textit{(bottom)} the distribution of true coefficients. 
        The $y$-axis in each panel shows the mean AUC based on 100 replications, and each line corresponds to one of six competing methods. 
        Overall, the EBGLM-based methods consistently achieve higher AUC in most settings, demonstrating robustness to varying settings. Additional plots are available in the Appendix \ref{appendix:plots}.
    }
    \label{fig:simulation}
\end{figure*}

The simulation results are presented in Figure \ref{fig:simulation}, where six methods are shown for enhanced clarity; additional plots are shown in Appendix \ref{appendix:plots}. Notably, the EBGLM-based methods consistently achieve higher AUC scores compared to other methods across nearly all scenarios. The EBGLM methods with different priors have similar performance, so for simplicity we report results only for EBGLM-pn in Figure \ref{fig:simulation}.

In experiments with varying sample sizes, all methods generally yield high and comparable AUC scores when the sample size exceeds 1500, though lasso and elastic net methods slightly underperform other methods. 
From the varying $p$ plot, we can see that, despite sparsity-inducing priors, the EBGLM methods perform well in low-dimensional settings, outperforming other methods. In comparison, lasso and elastic net methods show markedly worse AUC scores when  $p\geq 500$, and varbvs appears to be less effective in a dense scenario with $p=20$, where all coefficients are non-zero.

In the correlation plot, the noticeable decline in AUC suggests that high inter-feature correlation reduces predictive performance. Although all methods experience some degree of performance degradation under conditions of high correlation, EBGLM-based methods show greater resilience compared to others like L0Learn, MCP, and varbvs, which are more adversely affected.

In terms of sparsity level, AUC increases as the number of non-zero coefficients increases from $1$ to approximately $30$ and starts to decrease beyond this point. EBGLM-pn maintains superior performance, achieving the highest AUC up to a sparsity level of $50$. As the features become denser, lasso and SCAD begin to slightly outperform the other methods whereas L0Learn and varbvs tend to perform worse.

Lastly, all methods demonstrate comparable performance across Laplace and normal distributions of the coefficients, with minor AUC variations. When coefficients are generated from a uniform distribution, all methods show a significant decrease in AUC. This decline may be attributed to the reduced signal strength. When the coefficients are set at $1$, both the varbvs and L0Learn methods, along with the EBGLM, outperformed others, while the lasso and elastic net methods showed distinctly lower AUC values. Overall, the EBGLM methods maintained robustness across various distributions. 

\section{Real data Benchmark}

To demonstrate the performance of the proposed sparse EBGLM logistic regression on a real data example, we utilized the 20 Newsgroups dataset, which comprises a diverse collection of around 20k documents from 20 distinct groups and features high dimensionality ($p>n$). We preprocessed the data by eliminating headers, footers, and stopwords, and applied TF-IDF vectorization. We formulate the benchmark as binary classification tasks, specifically pairwise comparisons of categories from the same topic. The 20 groups are segmented into four main topics (recreational, computer-related, science-related, society and religion); within each topic, documents are further labelled with different categories and every category was compared against the others. For example, within science-related topic, there are four categories $\{$sci.crypt, sci.electronics, sci.med, sci.space$\}$ and the binary classification task is to predict one category vs the rest within the topic. Typically, the training dataset contains approximately $1000$ samples with $3000$ to $5000$ features, and the testing dataset comprises around $600$ samples. 

\begin{figure*}[ht]
    \centering
    \begin{subfigure}[b]{0.24\textwidth}
        \includegraphics[width=\columnwidth]{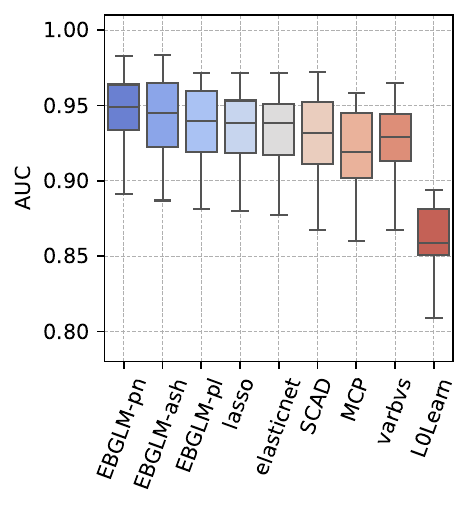}
        \caption{{Computer-related}}
    \end{subfigure}
        \begin{subfigure}[b]{0.24\textwidth}
        \includegraphics[width=\columnwidth]{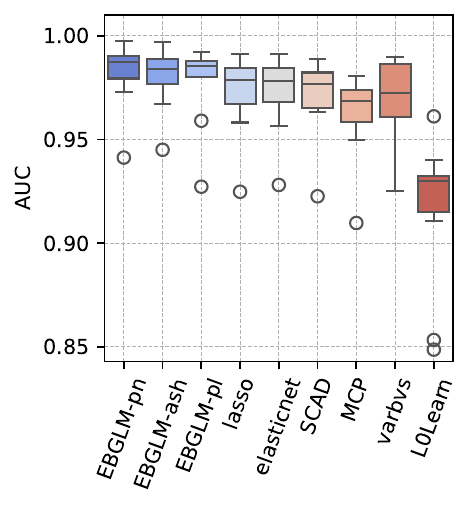}
        \caption{{Recreational }}
    \end{subfigure}
        \begin{subfigure}[b]{0.24\textwidth}
        \includegraphics[width=\columnwidth]{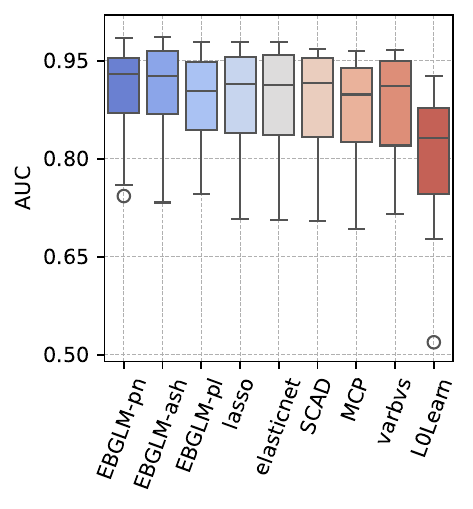}
        \caption{{Society and Religion }}
    \end{subfigure}
        \begin{subfigure}[b]{0.24\textwidth}
        \includegraphics[width=\columnwidth]{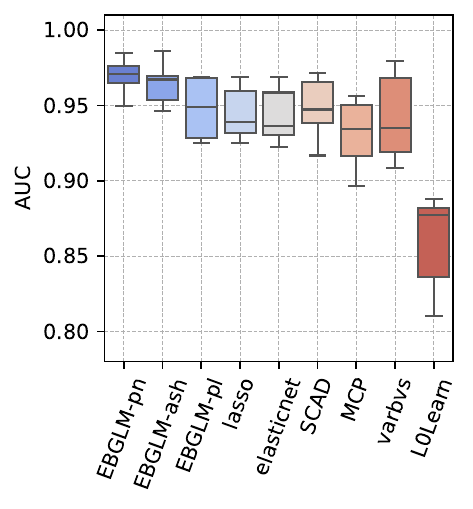}
        \caption{{Science-related}}
    \end{subfigure}
    \caption{\textbf{AUC Performance on the 20 Newsgroups Dataset}. 
        Each subfigure shows a box plot of AUC scores for different methods when classifying documents in one category versus the rest within that topic. 
        Each box plot summarizes performance across multiple random splits into train/test sets. 
        EBGLM-based methods yield higher median AUC and exhibit less variance (shorter boxes) across these four categories, suggesting both strong predictive performance and robustness in high-dimensional text classification.}
    \label{fig:news20}
\end{figure*}

Figure \ref{fig:news20} presents a box plot of the AUC scores for each method across all categories. The methods based on EBGLM consistently achieve higher AUC scores and exhibit less variability compared to others, as indicated by the generally shorter box heights. Apart from L0Learn, which underperforms as seen in our simulation studies, the performance of the other methods is relatively comparable. 

We also evaluated the proposed method on several publicly available classification datasets. In general, EBGLM demonstrates superior performance, as shown in Table \ref{table:real} in the Appendix \ref{appendix:uci}.

\section{Discussion and extensions}

In this paper, we introduced a novel variational inference framework for empirical Bayes generalized linear models, specifically applied to sparse logistic regression. Our approach is modular, allowing for the selection of the prior according to the researcher's preference and any exponential family distribution for the likelihood.
We relate the variational inference framework to optimization with penalty, and the corresponding objective function is analytical and a function of the posterior mean, supporting the use of various optimization methods.
The effectiveness of EBGLM has been demonstrated through simulation studies and real data benchmarks, where it consistently outperformed existing methods for sparse logistic regression. We chose logistic regression as a case study in this paper due to its popularity and widespread use, but the EBGLM framework can be directly used in other members of the exponential family.

While our focus is on using spike-and-slab priors for sparse GLM, alternative priors are also effective if the log marginal likelihood in the Bayesian normal means problem is computable, even requiring numerical methods. For example, the horseshoe prior lacks an analytic marginal likelihood expression, which can be resolved using numerical integration techniques like quadrature. A bottleneck in the optimization process is the root-finding task required to determine $z_g(\theta)$. This process can be optimized by using faster, derivative-based methods like Newton's method over the bisection method. As a final note, the proposed novel VI method is general and not limited to solve regression problems. It can be readily applied to other model settings such as matrix factorization, graphical models, and deep neural networks.

\newpage
\bibliographystyle{plainnat}
\bibliography{ebglm}

\newpage
\appendix
\section{Proofs}

\subsection{Proof of Lemma \ref{lemma:h}}
\label{appendix:lemma1_proof}

\begin{proof}
    We build an equivalence between $\tilde F(q,g)$ and $h(\bs\theta,g)$ by noting that
\begin{equation*}
\begin{split}
        \max_{q,g} \tilde F(q,g)&= \max_{\bs\theta,g}\max_{q:\E_{q}=\bs\theta} \tilde F(q,g),
        \\
        &=\max_{\bs\theta,g}-h(\bs\theta,g),
        \\
        &=-\min_{\bs\theta,g}h(\bs\theta,g),
\end{split}
\end{equation*}
where
\begin{equation*}
\begin{split}
    h(\bs\theta,g)  &:= \min_{q:\E_{q}=\bs\theta}-\tilde F(q,g),
    \\&= -\sum_i l(\bs x_i^T\bs\theta) + \min_{q:\E_{q}=\bs\theta} \sum_j \left(\frac{1}{2s^2_j(\bar{\bs\beta}_q)} V_{q_{\beta_j}} + D_{KL}(q_{\beta_j}||g)\right),
    \\&= -\sum_i l(\bs x_i^T\bs\theta) + \sum_j \min_{q_{j}:\E_{q_{j}}(\beta_j)=\theta_j}\left(\frac{1}{2s^2_j(\bs\theta)} V_{q_{\beta_j}} + D_{KL}(q_{\beta_j}||g)\right),
    \\
    &= -\sum_il(\bs x_i^T\bs\theta) + \sum_j r_j(\bs\theta,g).
\end{split}
\end{equation*}
\end{proof}

\subsection{Proof of Theorem \ref{thm:penalty}}
\label{appendix:thm1_proof}
\begin{proof}
Recall for $j=1,2,...,p$,
\[r_j( \bs\theta,g)=\min_{q_{j}:\E_{q_{j}}(\beta_j)=\theta_j}\left(\frac{1}{2s^2_j(\bs\theta)} V_{q_{j}} + D_{KL}(q_{j}||g)\right).\]
To solve for $r_j(\bs\theta,g)$, we formulate the Lagrangian multiplier as
\begin{equation}
    L(q_j,\lambda_0,\lambda_1) = \frac{1}{2s_j^2(\bs\theta)}\int q_j(\beta_j-\theta_j)^2 + D_{KL}(q_j||g) + \lambda_0(\int q_j-1) + \lambda_1(\int \beta_j q_j -\theta_j),
\end{equation}
where the last two terms come from the constraints that $q_{\beta_j}$ is a density and its mean is $\theta_j$. Taking derivative of $L(q_j,\lambda_0,\lambda_1)$ with respect to $q_j$ and set it to 0, we have
\begin{equation*}
\begin{split}
        &-\log g(\beta_j) +1+\log q_j + \lambda_0 + \lambda_1\beta_j 
        +\frac{1}{2s_j^2(\bs\theta)}(\beta_j-\theta_j)^2= 0
        \\
        &\Longrightarrow q_j = g(\beta_j)e^{-\frac{1}{2s_j^2(\bs\theta)}(\beta_j-\theta_j)^2 - \lambda_1 \beta_j- \lambda_0-1}.
\end{split}
\end{equation*}
By completing the square, the optimal $q_j$ thus has the form 
\begin{equation*}
    q_j = \frac{g(\beta_j)N(z_g(\theta_j);\beta_j,s_j^2(\bs\theta))}{c(z_g(\theta_j))},
\end{equation*}
where $c(z_g(\theta_j)) = \int g(b)N(z_g(\theta_j);b,s_j^2(\bs\theta)) db$ and $z_g(\theta_j)$ is the ``observation'' that leads to the posterior mean $\theta_j$ in a Bayesian normal mean problem with prior $g(\cdot)$ and variance $s_j^2(\bs\theta)$.  

The $r_j(\bs\theta,g)$ now evaluates as 
\begin{equation*}
\begin{split}
        r_j(\bs\theta,g) &= -\log c(z_g(\theta_j)) -\E_q\frac{(z_g(\theta_j)-\theta_j + \theta_j - \beta_j)^2}{2s_j^2(\bs\theta)} - \frac{1}{2}\log 2\pi s_j^2(\bs\theta) + \frac{1}{2s^2}V_{q_j}
        \\
        &= -l_{\text{NM}}(z_g(\theta_j);g,s_j^2(\bs\theta))-\frac{(z_g(\theta_j)-\theta_j)^2}{2s_j^2(\bs\theta)} - \frac{1}{2}\log 2\pi s_j^2(\bs\theta)
        \\
        &= -l_{\text{NM}}(z_g(\theta_j);g,s_j^2(\bs\theta))+\log N(z_g(\theta_j);\theta_j,s_j^2(\bs\theta)).
\end{split}
\end{equation*}

A remaining question is how to determine $z_g(\theta_j)$, which requires solving an ``inverse'' BNM problem: finding the observations that result in a given posterior mean and prior. In fact, $z_g(\theta_j)$ is an implicit function of $\theta_j$. According to Tweedie's formula \cite{efron2011tweedie}, $z_g(\theta_j)$ is found by solving for $z$ in the following root finding problem for each $j$:
\begin{equation*}
    \theta_j = z + s_j^2(\bs\theta)l'_{\text{NM}}(z;g,s_j^2(\bs\theta)),
\end{equation*}
where $l'_{\text{NM}}(\cdot)$ is the derivative of $l_{\text{NM}}(\cdot)$ with respect to $z$. We employ the bisection method to solve this problem due to its stability.

\end{proof}

\section{Model details}

\subsection{Sparse Gaussian linear regression}

For Gaussian distribution, we have $\phi=\sigma^2$ and $b(\mu) = \mu^2/2$. The objective function is then 
\begin{equation*}
    h = -y^TX\bs\theta/\sigma^2 + \sum_i\frac{(x_i^T\bs\theta)^2}{2\sigma^2}  + \frac{n}{2}\log2\pi\sigma^2 + \sum_i\frac{y_i^2}{2\sigma^2} + \sum_j r_j(\bs\theta,g),
\end{equation*}
where \[s_j^2(\bs\theta) = \sigma^2\left(\sum_i x_{ij}^2\right)^{-1}\]

\subsection{Sparse Poisson regression}

For Poisson distribution, we have $\phi=1$ and $b(\mu) = \exp(\mu)$. The objective function is then 
\begin{equation*}
    h = -y^TX\bs\theta + \sum_i\exp(x_i^T\bs\theta)  + \sum_j r_j( \bs\theta,g),
\end{equation*}
where \[s_j^2(\bs\theta) = \left(\sum_i\exp(x_i^T\bs\theta)  x_{ij}^2\right)^{-1}\]

\subsection{Sparse Logistic regression}

For Bernoulli distribution, we have $\phi=1$ and $b(\mu) = \log(1+\exp(\mu))$. The objective function is then 
\begin{equation*}
    h = -y^TX\bs\theta + \sum_i\log(1+\exp(x_i^T\bs\theta)) + \sum_j r_j(\bs\theta,g),
\end{equation*}
where \[s_j^2(\bs\theta) = \left(\sum_i p_i(1 - p_i) x_{ij}^2\right)^{-1}, p_i = \frac{\exp(x_i^T\bs\theta)}{1+\exp(x_i^T\bs\theta)}\]

\subsection{Point Laplace prior marginal likelihood}
\label{appendix:point_laplace}

With Point-Laplace prior, the BNM problem is
\begin{equation*}
    \begin{split}
        &x|\mu\sim N(\mu,s^2)
        \\
        &\mu\sim \pi_0\delta_0+(1-\pi_0)\text{Lap}(0,b)
    \end{split}
\end{equation*}

Marginal likelihood:
\begin{equation*}
    \begin{split}
        L(x;\pi_0,b) &= \pi_0 N(x;0,s^2) + (1-\pi_0) \int_{-\infty}^{\infty} N(x;\mu,s^2)\frac{1}{2b}\exp(-|\mu|/b)d\mu
        \\
        &=\pi_0 N(x;0,s^2) + (1-\pi_0) \int_{0}^{\infty} N(x;\mu,s^2)\frac{1}{2b}\exp(-\mu/b)d\mu \\&\qquad + (1-\pi_0) \int_{-\infty}^{0} N(x;\mu,s^2)\frac{1}{2b}\exp(\mu/b)d\mu
        \\
        &=\pi_0 N(x;0,s^2) + (1-\pi_0) \frac{1}{4b}\exp(\frac{s^2-2bx}{2b^2})\text{erfc}(\frac{s^2-bx}{\sqrt{2s^2}b}) \\&\qquad + (1-\pi_0) \frac{1}{4b}\exp(\frac{s^2+2bx}{2b^2})\text{erfc}(\frac{s^2+bx}{\sqrt{2s^2}b}),
    \end{split}
\end{equation*}
where $\text{erfc}(z) = 1- \text{erf}(z) = \frac{2}{\sqrt{\pi}}\int_z^{\infty}\exp(-t^2)dt$.

Derivative of log marginal likelihood $l(x;\pi_0,b) = \log L(x;\pi_0,b)$ w.r.t $x$ is 

\begin{equation*}
    \begin{split}
        \frac{d l(x)}{dx} &= \frac{L'(x)}{L(x)},
    \end{split}
\end{equation*}
where 
\begin{equation*}
    \begin{split}
        L'(x) &= \pi_0 N(x;0,s^2)\times (-x/s^2)  \\&\qquad + (1-\pi_0)\left(\frac{\exp \left(\frac{s^2-2 b x}{2 b^2}-\frac{\left(s^2-b x\right)^2}{2 b^2 s^2}\right)}{2 \sqrt{2 \pi s^2} b}-\frac{e^{\left(s^2-2 b x\right) /\left(2 b^2\right)} \text{erfc}\left(\frac{s^2-b x}{\sqrt{2 s^2} b}\right)}{4 b^2}\right) \\&\qquad + (1-\pi_0)\left( \frac{e^{\left(2 b x+s^2\right) /\left(2 b^2\right)} \text{erfc}\left(\frac{b x+s^2}{\sqrt{2s^2} b}\right)}{4 b^2}-\frac{\exp \left(\frac{2 b x+s^2}{2 b^2}-\frac{\left(b x+s^2\right)^2}{2 b^2 s^2}\right)}{2 \sqrt{2 \pi s^2} b}\right)
    \end{split}
\end{equation*}

\section{Additional results}

\subsection{An illustration of the variational posterior distribution from the VI}
\label{sec:illustration_vi}

We evaluate the accuracy of the proposed VI algorithm on simulated data using logistic regression. Specifically, we simulate $p = 10$ regression coefficients from independent standard normal priors and generate $n = 1000$ observations from the likelihood. We consider two scenarios for the feature matrix $X$, with correlation levels $\rho \in {0, 0.8}$. We compare the posterior distributions obtained via our VI method with those computed using MCMC (implemented in \texttt{PyMC}, with 1000 burn-in and 3000 posterior samples).

\begin{figure*}[ht]
    \centering
    \begin{subfigure}[b]{0.95\textwidth}
        \includegraphics[width=\columnwidth]{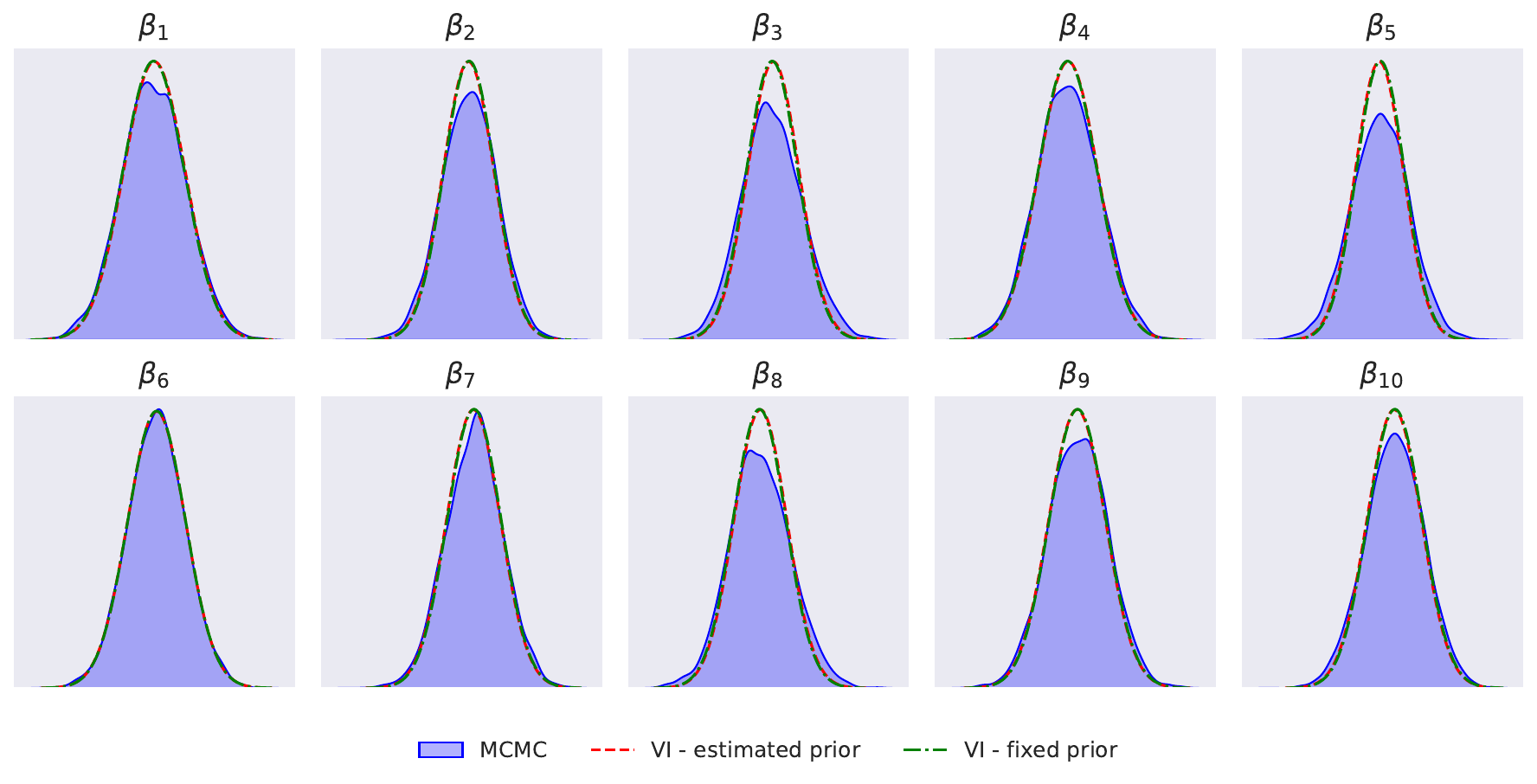}
        \caption{Correlations among features $\rho=0$}
        \label{fig:post_dist_rho0_normal}
    \end{subfigure}
    \begin{subfigure}[b]{0.95\textwidth}
        \includegraphics[width=\columnwidth]{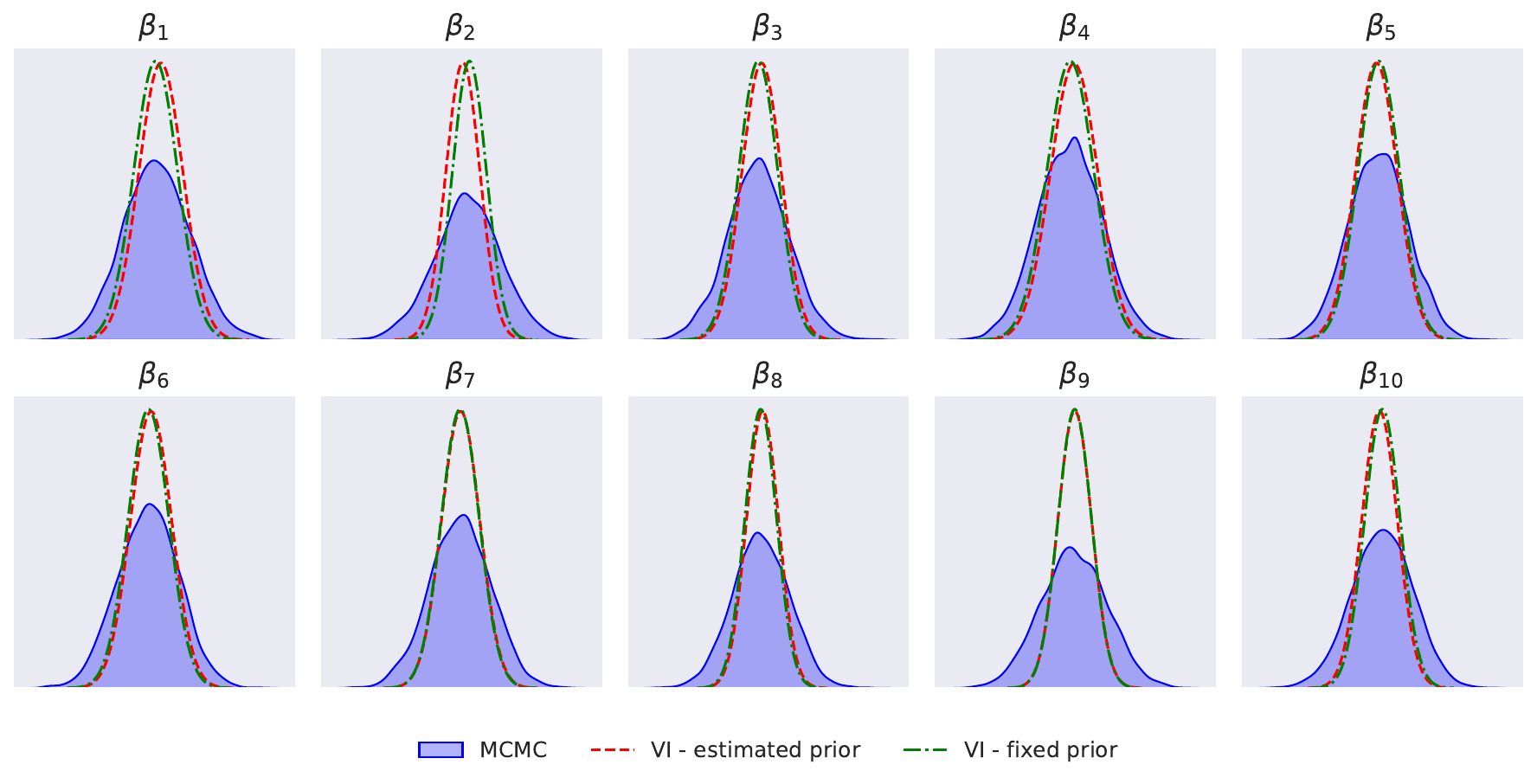}
        \caption{Correlations among features $\rho=0.8$}
        \label{fig:post_dist_rho08_normal}
    \end{subfigure}
    \caption{Comparison of marginal variational posterior densities of regression coefficients in Bayesian logistic regression. Regression coefficients are drawn from standard normal prior $\beta_j\sim N(0,1)$ for $j=1,2,...,10.$ }
\end{figure*}

As shown in Figure~\ref{fig:post_dist_rho0_normal}, when the features are weakly correlated, the mean-field VI (MFVI) closely approximates both the posterior distribution and the posterior mean. In contrast, when the features are highly correlated (Figure~\ref{fig:post_dist_rho08_normal}), MFVI tends to underestimate posterior uncertainty—a well-known phenomenon—but still provides accurate estimates of the posterior mean. We also compare VI with an estimated prior (the empirical Bayes) to VI with a fixed true prior (standard normal). The resulting variational posteriors are nearly identical, suggesting that the estimated prior effectively captures the true prior distribution.

\begin{figure*}[ht]
    \centering
    \begin{subfigure}[b]{0.95\textwidth}
        \includegraphics[width=\columnwidth]{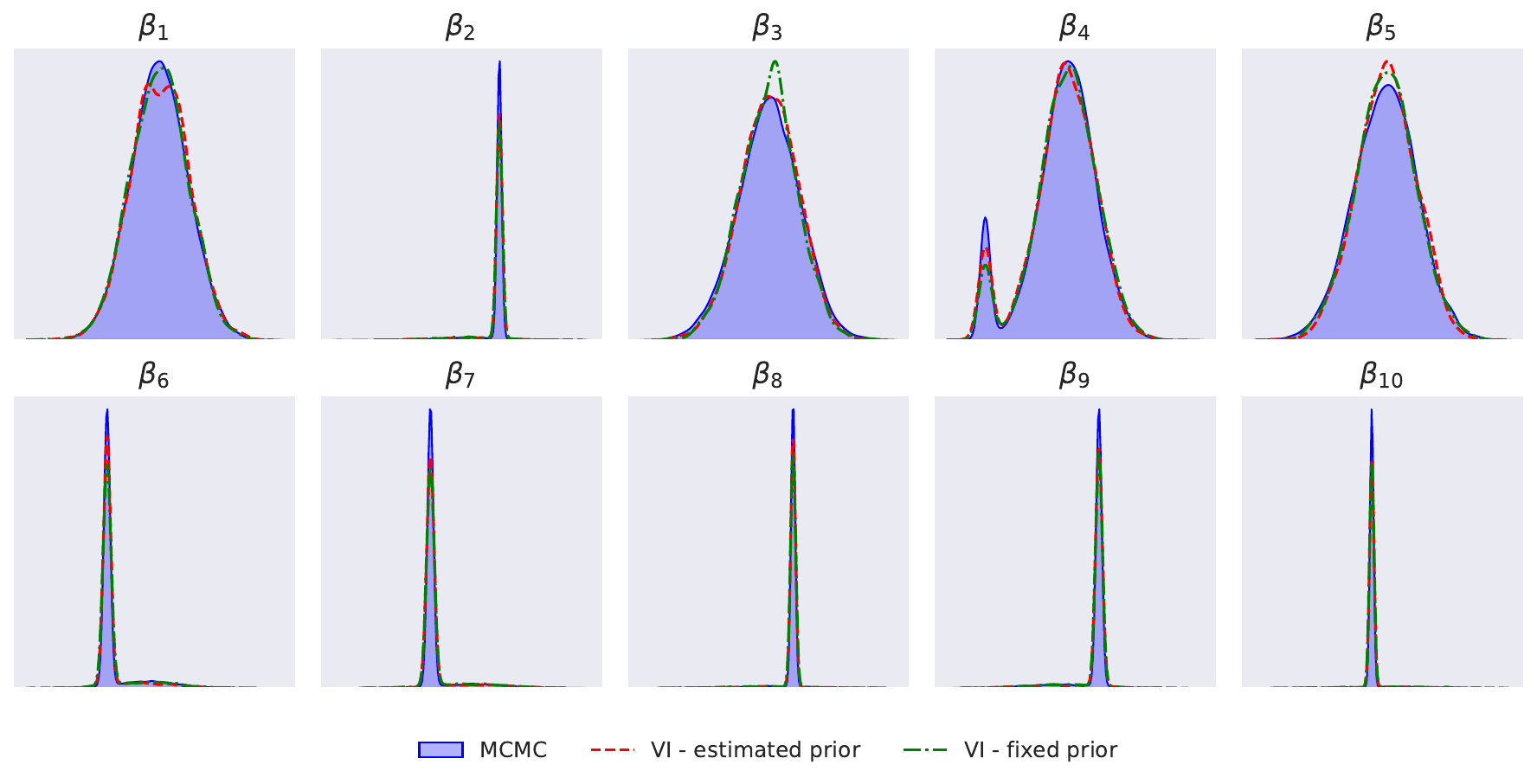}
        \caption{Marginal variational posterior densities.}
        \label{fig:post_dist_rho0_point_normal}
    \end{subfigure}
    \begin{subfigure}[b]{0.95\textwidth}
        \includegraphics[width=\columnwidth]{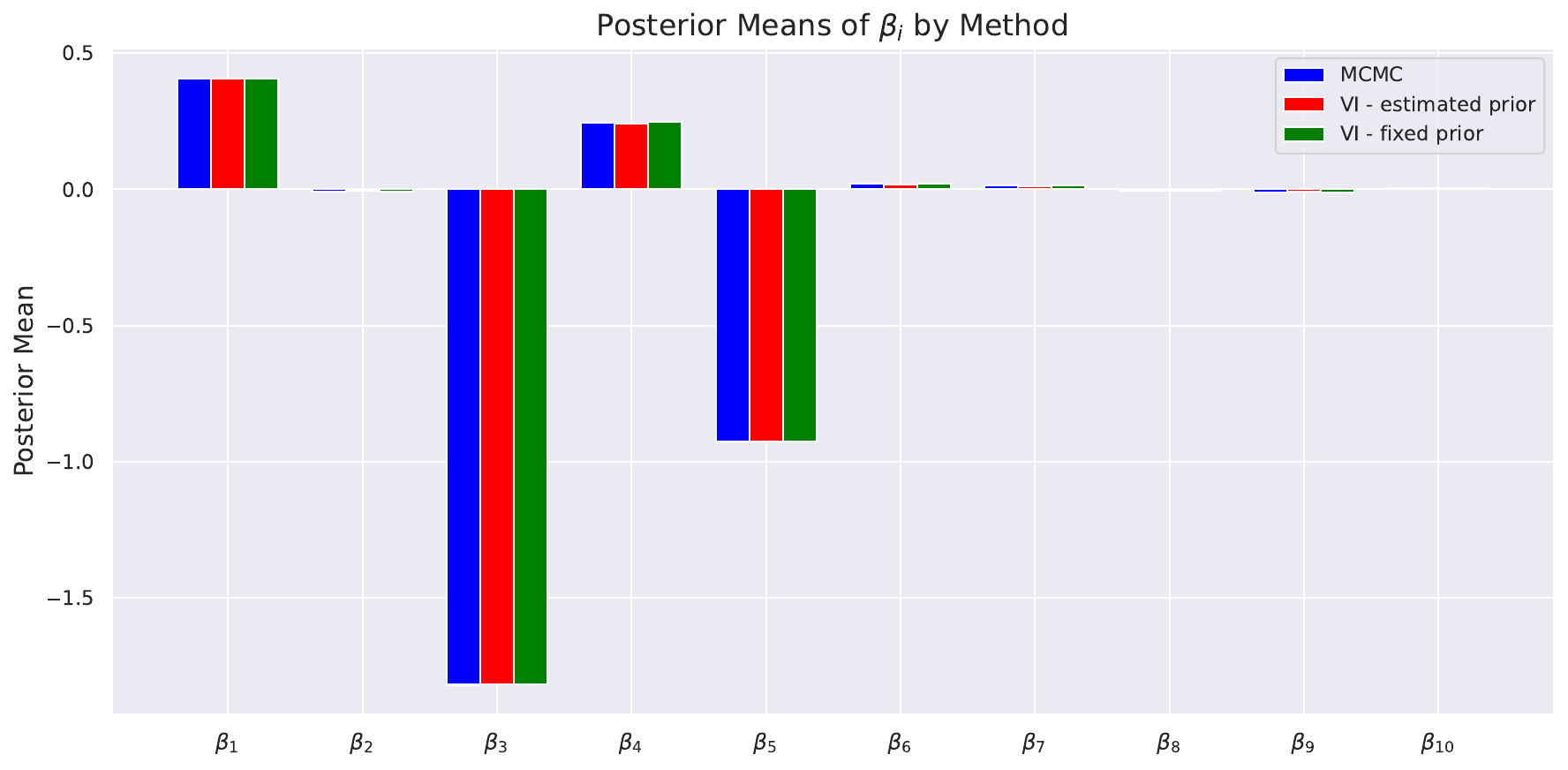}
        \caption{Bar plot of posterior mean.}
        \label{fig:post_mean_rho0_point_normal}
    \end{subfigure}
    \caption{Comparison of marginal variational posterior densities and posterior mean of regression coefficients in Bayesian logistic regression. Regression coefficients are drawn from point normal prior $\beta_j\sim 0.5\delta_0+0.5N(0,1)$ for $j=1,2,...,10.$ Correlations among features $\rho=0$. }
\end{figure*}

\begin{figure*}[ht]
    \centering
    \begin{subfigure}[b]{0.95\textwidth}
        \includegraphics[width=\columnwidth]{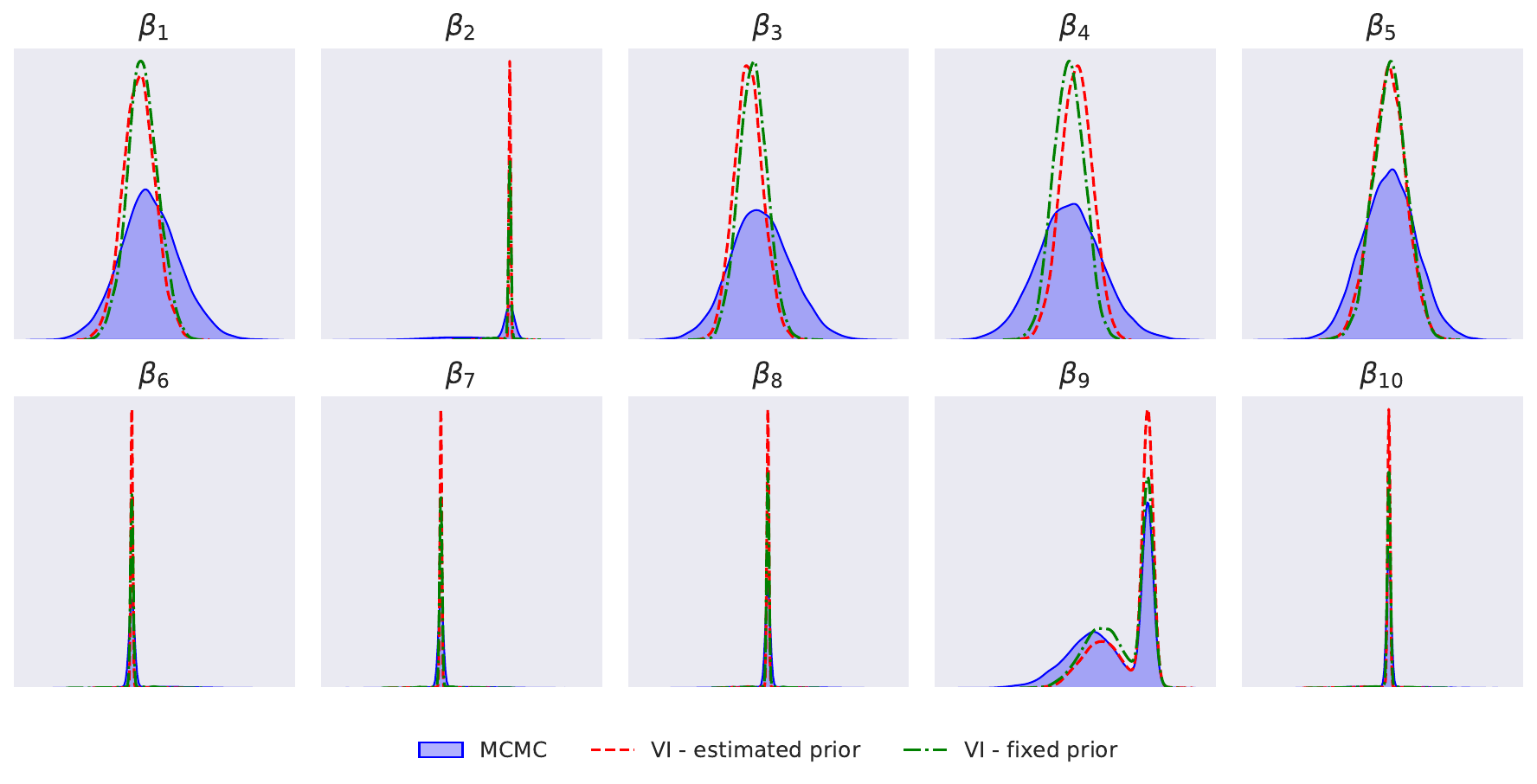}
        \caption{Marginal variational posterior densities.}
        \label{fig:post_dist_rho08_point_normal}
    \end{subfigure}
    \begin{subfigure}[b]{0.95\textwidth}
        \includegraphics[width=\columnwidth]{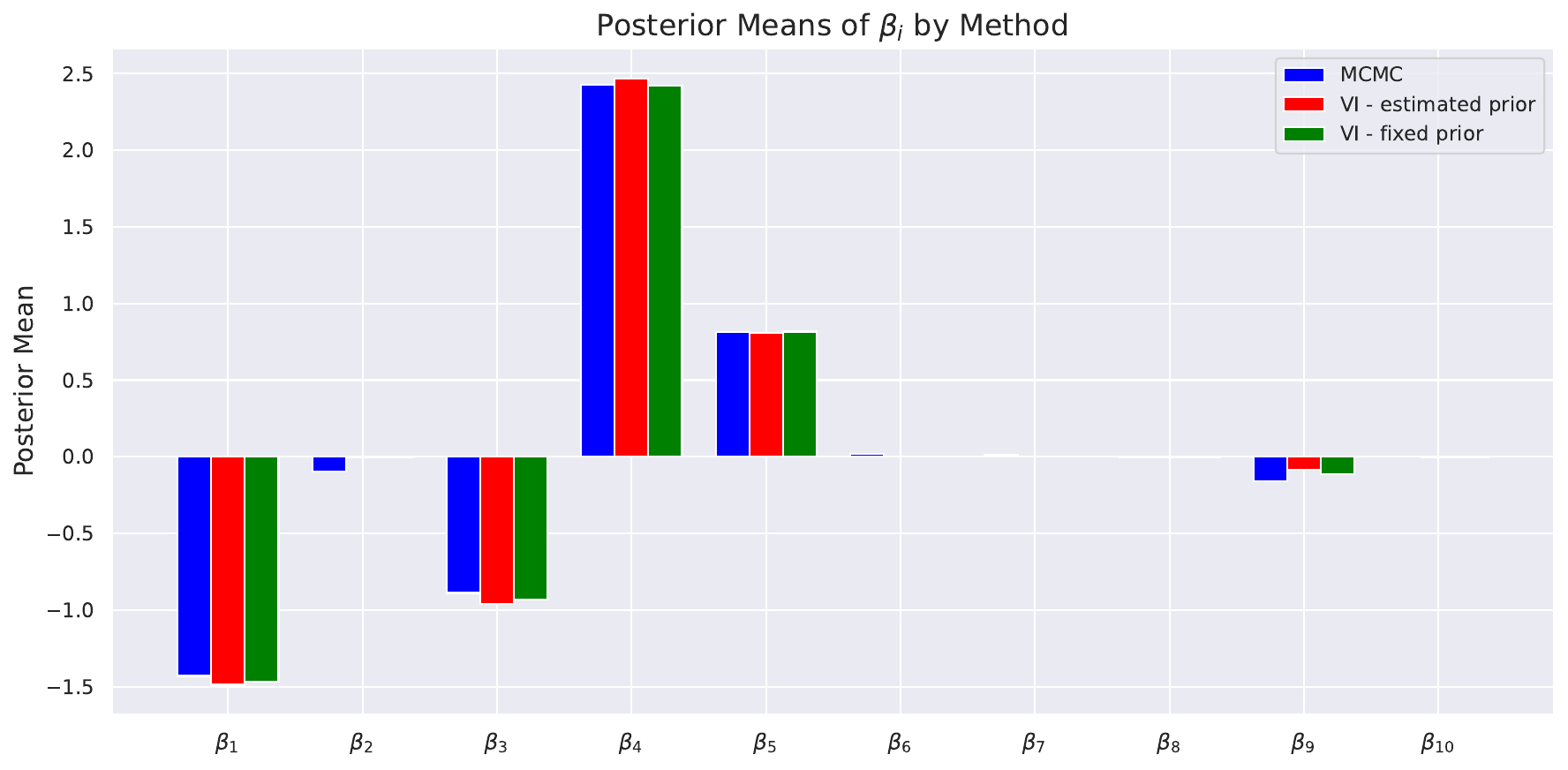}
        \caption{Bar plot of posterior mean.}
        \label{fig:post_mean_rho08_point_normal}
    \end{subfigure}
    \caption{Comparison of marginal variational posterior densities and posterior mean of regression coefficients in Bayesian logistic regression. Regression coefficients are drawn from point normal prior $\beta_j\sim 0.5\delta_0+0.5N(0,1)$ for $j=1,2,...,10.$ Correlations among features $\rho=0.8$. }
\end{figure*}

In a second experiment, we use the same simulation setup but change the prior on the regression coefficients to a spike-and-slab prior: $\beta_j \sim 0.5\delta_0 + 0.5\mathcal{N}(0,1)$ for $j = 1, \ldots, 10$. The conclusions remain consistent: as shown in Figures~\ref{fig:post_dist_rho0_point_normal} and \ref{fig:post_dist_rho08_point_normal}, our VI method captures the posterior mean well. This is further illustrated in Figures~\ref{fig:post_mean_rho0_point_normal} and \ref{fig:post_mean_rho08_point_normal}, where the estimated posterior means closely match those from MCMC, regardless of feature correlation.

\newpage
\subsection{UCI data benchmark}
\label{appendix:uci}
In this study, we evaluate the performance of the proposed sparse EBGLM logistic regression method on several publicly available classification datasets. Specifically, we conduct experiments on six datasets from the UCI ML dataset collection \cite{kelly2023uci}: Adult, Annealing, Ionosphere, Heart Disease, Abalone, and Pediatric. 
For the evaluation, we adopt a randomized split approach, dividing the data into $60\%$ for training and $40\%$ for testing, and we repeat this process 20 times per dataset. In these datasets, the sample size exceeds the number of features ($n > p$). The performance metrics (averaged AUC, $F_1$ score, and accuracy) are reported in Table \ref{table:real} for each method across all repetitions and datasets, using a 0.5 cutoff for the $F_1$ and accuracy calculations. Overall, while the performance of all methods is comparable, the EBGLM approaches demonstrate marginally superior performance.

\begin{table}[t]

\centering
\caption{UCI ML datasets benchmark: model metrics averaged across all repetitions and datasets.}
\begin{tabular}{lccc}
\toprule
Model      & AUC                & $F_1$    & Accuracy \\
\midrule
EBGLM-pn   & \textbf{0.921} (0.043) & \textbf{0.823} & \textbf{0.867} \\
EBGLM-pl   & \textbf{0.921} (0.044) & 0.821 & \textbf{0.867} \\
EBGLM-ash  & \textbf{0.921} (0.043) & 0.822 & 0.866 \\
lasso      & 0.920 (0.045)         & 0.814 & 0.864 \\
varbvs     & 0.918 (0.044)         & 0.818 & 0.862 \\
MCP        & 0.912 (0.040)         & 0.816 & 0.860 \\
SCAD       & 0.915 (0.039)         & 0.817 & 0.861 \\
L0Learn    & 0.908 (0.057)         & 0.820 & 0.865 \\
\bottomrule
\end{tabular}
\label{table:real}
\end{table}

\subsection{Additional plots}
\label{appendix:plots}

\begin{figure*}[ht]
    \centering
    \begin{subfigure}[b]{0.4\textwidth}
        \includegraphics[width=\columnwidth]{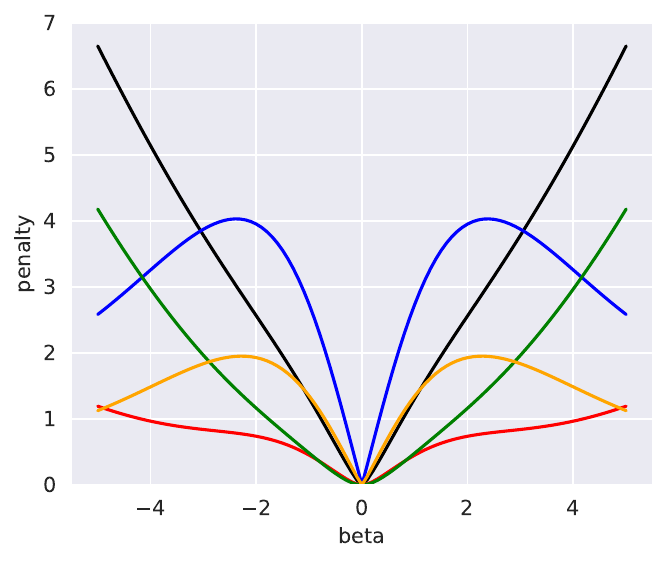}
        \caption{Point Normal prior penalty functions with different combination of $\pi_0$ and $\sigma^2$ in logistic regression.}
        \label{fig:pn_penalty}
    \end{subfigure}
    \hspace{0.4in}
    \begin{subfigure}[b]{0.4\textwidth}
        \includegraphics[width=\columnwidth]{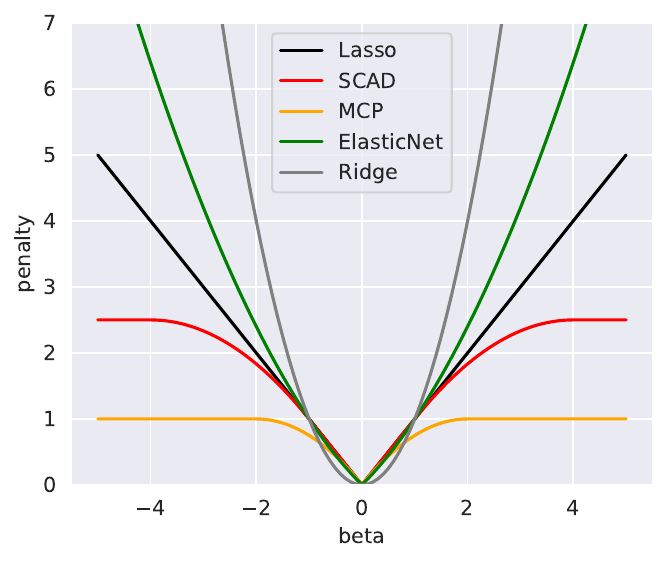}
        \caption{Other popular penalty functions that have been widely used in penalized regression methods.}
        \label{fig:other_penalty}
    \end{subfigure}
    \caption{\textbf{Comparison of penalty functions}. \textit{(Left)} Penalty curves for the Point Normal prior under different combinations of the mixing probability \(\pi_0\) and variance \(\sigma^2\), shown here in a logistic regression setting. These curves illustrate how different hyperparameter choices control the amount of shrinkage on the coefficients. 
        \textit{(Right)} Penalty curves from several well-known regularizers commonly used in penalized regression. 
        The $x$-axis represents the coefficient value, and the $y$-axis represents the corresponding penalty.}
    \label{fig:penalty}
\end{figure*}



\begin{figure}
    \centering
    \includegraphics[width=0.7\textwidth]{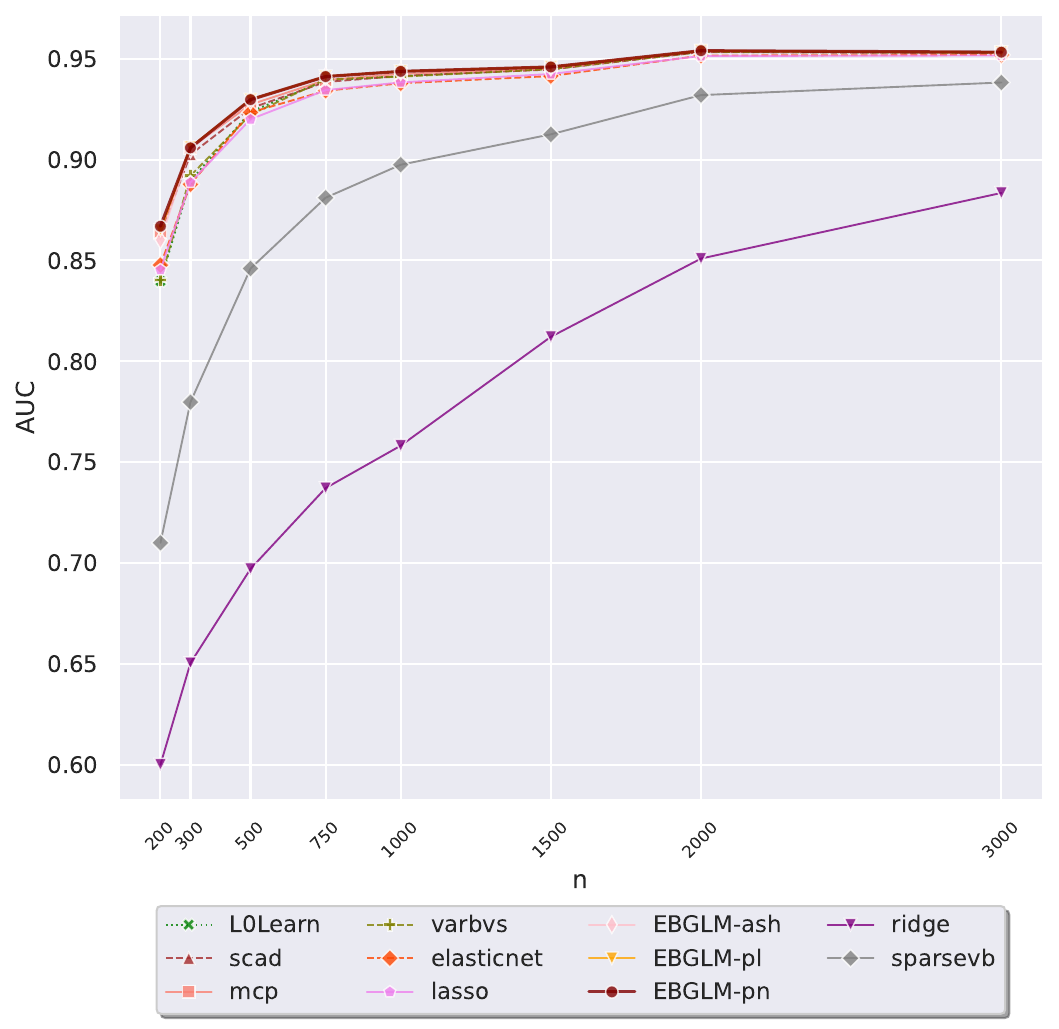}
    \caption{Simulation: varying $n$}
    \label{fig:simu_n_app}
\end{figure}

\begin{figure}
    \centering
    \includegraphics[width=0.7\textwidth]{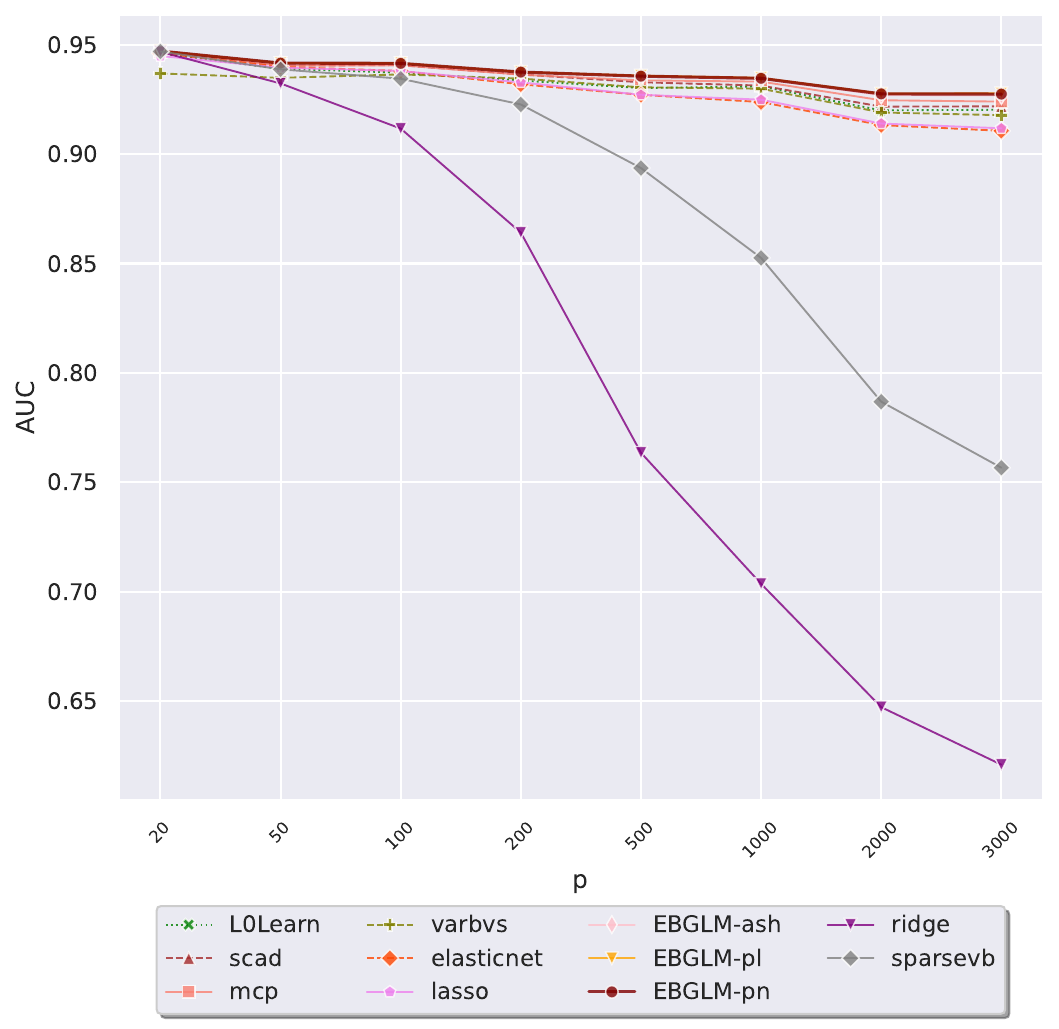}
    \caption{Simulation: varying $p$}
    \label{fig:simu_p_app}
\end{figure}

\begin{figure}
    \centering
    \includegraphics[width=0.7\textwidth]{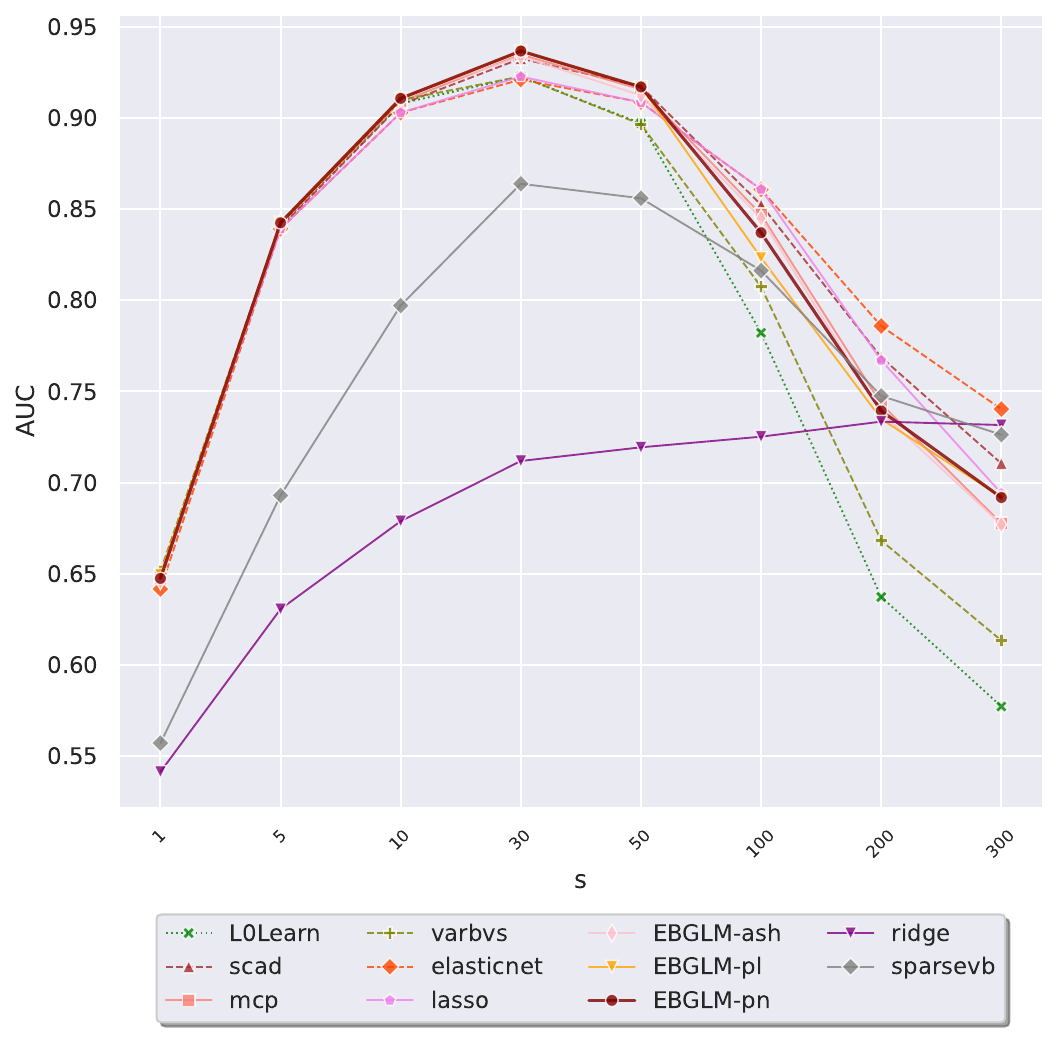}
    \caption{Simulation: varying $s$}
    \label{fig:simu_s_app}
\end{figure}

\begin{figure}
    \centering
    \includegraphics[width=0.8\textwidth]{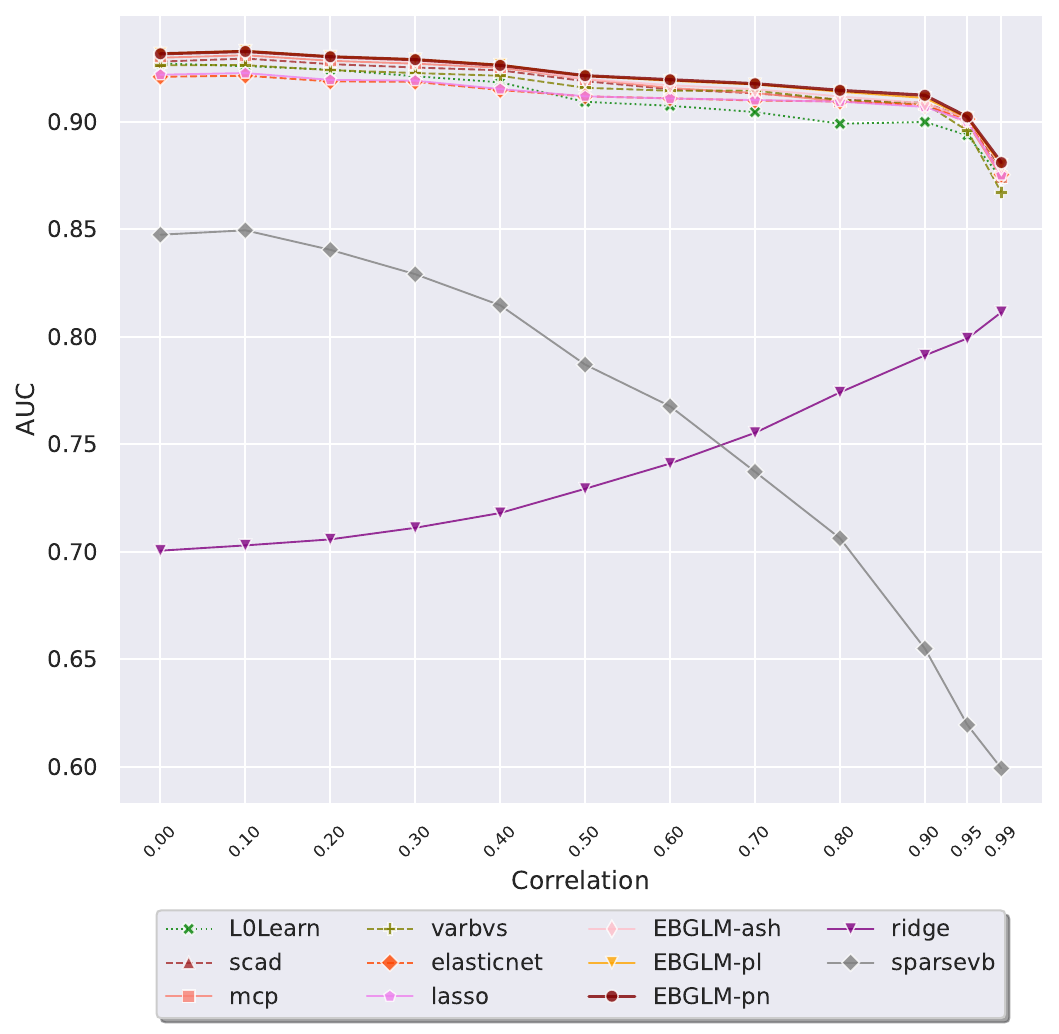}
    \caption{Simulation: varying $r$}
    \label{fig:simu_r_app}
\end{figure}

\begin{figure}
    \centering
    \includegraphics[width=0.8\textwidth]{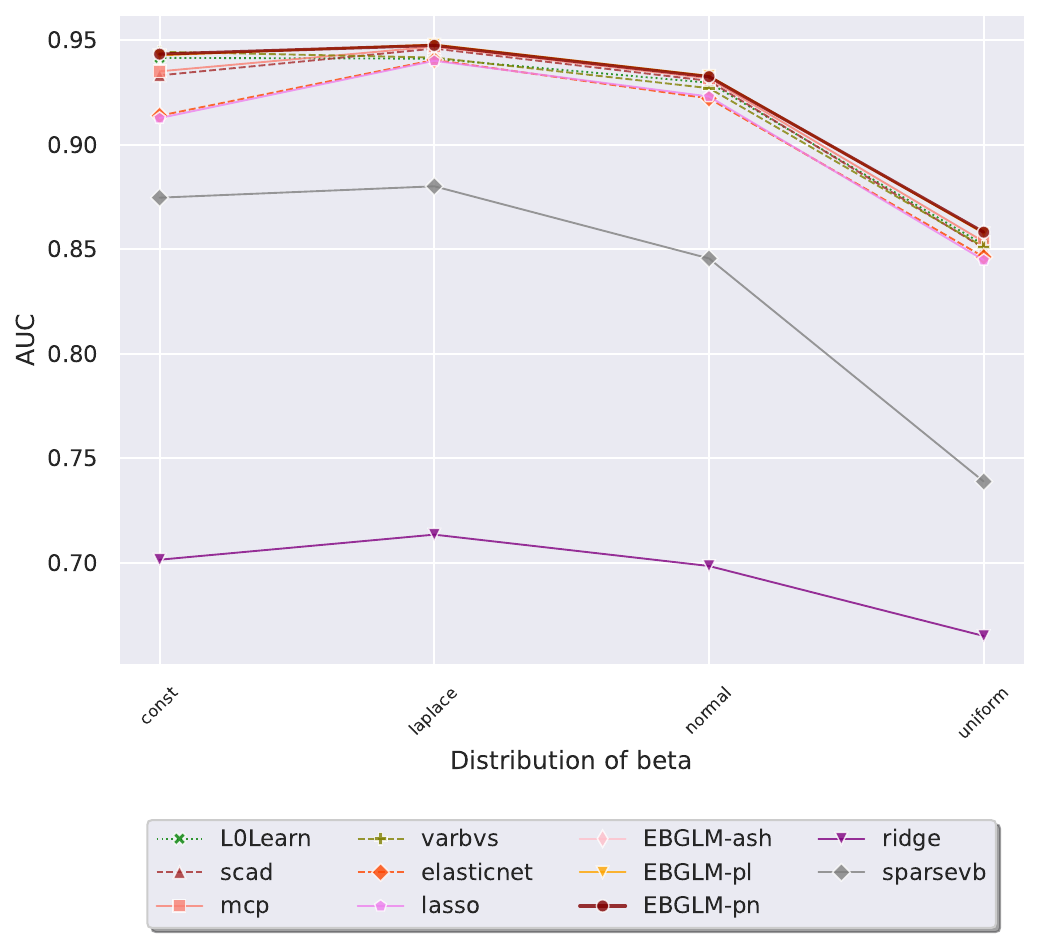}
    \caption{Simulation: varying beta distributions}
    \label{fig:simu_beta_dist_app}
\end{figure}

\end{document}